\pdfoutput=1
\documentclass[11pt]{article}

\usepackage{EMNLP2023}

\usepackage{times}
\usepackage{latexsym}
\usepackage{booktabs}

\usepackage[T1]{fontenc}

\usepackage[utf8]{inputenc}
\usepackage{titlesec}
\titlespacing*{\title}{0pt}{\baselineskip}{\baselineskip}

\usepackage{microtype}
\usepackage{colortbl}
\usepackage{enumitem}
\usepackage{arabtex}
\usepackage{utf8}
\setcode{utf8}
\usepackage{hyperref}
\usepackage{fdsymbol}
\usepackage{graphicx}
\usepackage{pgf}
\usepackage{subcaption}
\usepackage{svg}
\usepackage{caption}
\usepackage{enumitem}
\usepackage{multirow}
\usepackage{multicol}
\usepackage{xspace}
\usepackage{comment}
\usepackage{array}
\usepackage{booktabs}

\newcommand{\textclass}[1]{\text{#1}\xspace}
\newcommand{\hw}{\textclass{Human-Written}}
\newcommand{\mg}{\textclass{Machine-Generated}}
\newcommand{\mwmh}{\textclass{Machine-Written Machine-Humanized}}
\newcommand{\hwmp}{\textclass{Human-Written Machine-Polished}}

\title{LLM-DetectAIve: a Tool for Fine-Grained\\Machine-Generated Text Detection}

\author{
 \textbf{Mervat Abassy,\textsuperscript{1,2}$^*$}
 \textbf{Kareem Elozeiri,\textsuperscript{1,3}$^*$}
 \textbf{Alexander Aziz,\textsuperscript{1,4}$^*$}
 \textbf{Minh Ngoc Ta,\textsuperscript{1,5}$^*$}\\
 \textbf{Raj Vardhan Tomar,\textsuperscript{1,6}$^*$}
 \textbf{Bimarsha Adhikari,\textsuperscript{1,9}$^*$}
 \textbf{Saad El Dine Ahmed,\textsuperscript{1,2}$^*$} \\
 \textbf{Yuxia Wang,\textsuperscript{1}}
 \textbf{Osama Mohammed Afzal,\textsuperscript{1}}
 \textbf{Zhuohan Xie,\textsuperscript{1}}
 \textbf{Jonibek Mansurov,\textsuperscript{1}}\\
 \textbf{Ekaterina Artemova,\textsuperscript{7}}
 \textbf{Vladislav Mikhailov,\textsuperscript{8}}
 \textbf{Rui Xing,\textsuperscript{1}}
 \textbf{Jiahui Geng,\textsuperscript{1}} 
 \textbf{Hasan Iqbal,\textsuperscript{1}} \\
 \textbf{Zain Muhammad Mujahid,\textsuperscript{1}}
 \textbf{Tarek Mahmoud,\textsuperscript{1}}
 \textbf{Akim Tsvigun,\textsuperscript{10}}
 \textbf{Alham Fikri Aji,\textsuperscript{1}} \\ 
 \textbf{Artem Shelmanov,\textsuperscript{1}}
 \textbf{Nizar Habash,\textsuperscript{1,9}}
 \textbf{Iryna Gurevych,\textsuperscript{1}}
 \textbf{Preslav Nakov\textsuperscript{1}} \\
 \\
 \textsuperscript{1}MBZUAI,
 \textsuperscript{2}Alexandria University,
 \textsuperscript{3}Zewail City of Science and Technology, \\
 \textsuperscript{4}University of Florida,
 \textsuperscript{5}Hanoi University of Science and Technology, \\
 \textsuperscript{6}Cluster Innovation Center, University of Delhi, \textsuperscript{7}Toloka AI, \textsuperscript{8}University of Oslo,  \\
 \textsuperscript{9}New York University Abu Dhabi, \textsuperscript{10}KU Leuven \\
}

\begin{document}
\maketitle
\def\thefootnote{*}\footnotetext{Equal contribution.}\def\thefootnote{\arabic{footnote}}
\vspace{3cm}

\begin{abstract}
 The ease of access to large language models (LLMs) has enabled a widespread of machine-generated texts, and now it is often hard to tell whether a piece of text was human-written or machine-generated. This raises concerns about potential misuse, particularly within educational and academic domains. Thus, it is important to develop practical systems that can automate the process. Here, we present one such system, \textbf{LLM-DetectAIve}, designed for fine-grained detection. Unlike most previous work on machine-generated text detection, which focused on binary classification, LLM-DetectAIve supports four categories: (i)~human-written, (ii)~machine-generated, (iii)~machine-written, then machine-humanized, and (iv)~human-written, then machine-polished. Category (iii) aims to detect attempts to obfuscate the fact that a text was machine-generated, while category (iv) looks for cases where the LLM was used to polish a human-written text, which is typically acceptable in academic writing, but not in education. Our experiments show that LLM-DetectAIve can effectively identify the above four categories, which makes it a potentially useful tool in education, academia, and other domains.
 LLM-DetectAIve is publicly accessible at \url{https://github.com/mbzuai-nlp/LLM-DetectAIve}.\footnote{This work was done during a summer internship at the NLP department, MBZUAI.} 
 The video describing our system is available at \url{https://youtu.be/E8eT_bE7k8c}.

\end{list} 

\end{abstract}

\section{Introduction}





The development of advanced large language models (LLMs), such as GPT-4, Claude-3.5, Gemini-1.5, Llama-70b~\citep{gpt4, claude3, Gemini2023, dubey2024llama3herdmodels}, improved the prevalence and the coherence of machine-generated content. This trend makes it increasingly difficult to differentiate between texts produced by machines from such written by humans~\cite{macko2023multitude,wang2024m4gtbenchevaluationbenchmarkblackbox,wang-etal-2024-m4}. As a result, there have been growing concerns about the authenticity and integrity of textual content \cite{crothers2023machine,tang2024science}.

While many detectors have been developed to address this new challenge \cite{pmlr-v202-mitchell23a,wang2024semeval}, they often struggle to keep up with the rapid development of LLMs. Generations produced by new models are hard to detect as they become more coherent and represent out-of-distribution instances, compared to what detecting systems saw during training~\cite{macko2024authorship,koike2024outfox}. 
Moreover, the use of prompting to generate more human-like texts or applying LLMs to refine or change the tone of human writings further complicates detection.

\begin{figure*}[t]
    \centering
    \includegraphics[scale=0.25,trim=0.8cm 0 0 0]{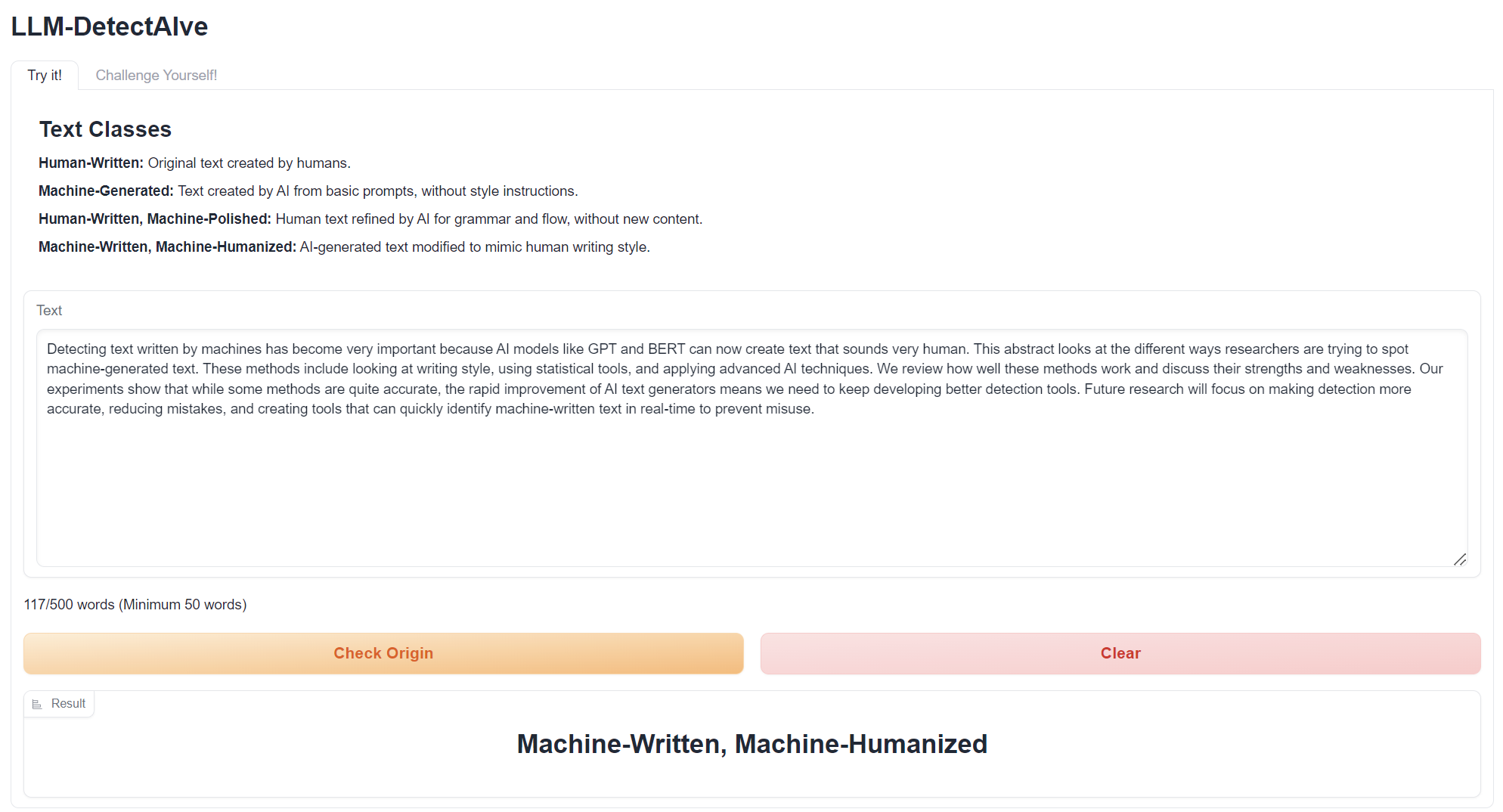} 
    \includegraphics[scale=0.25]{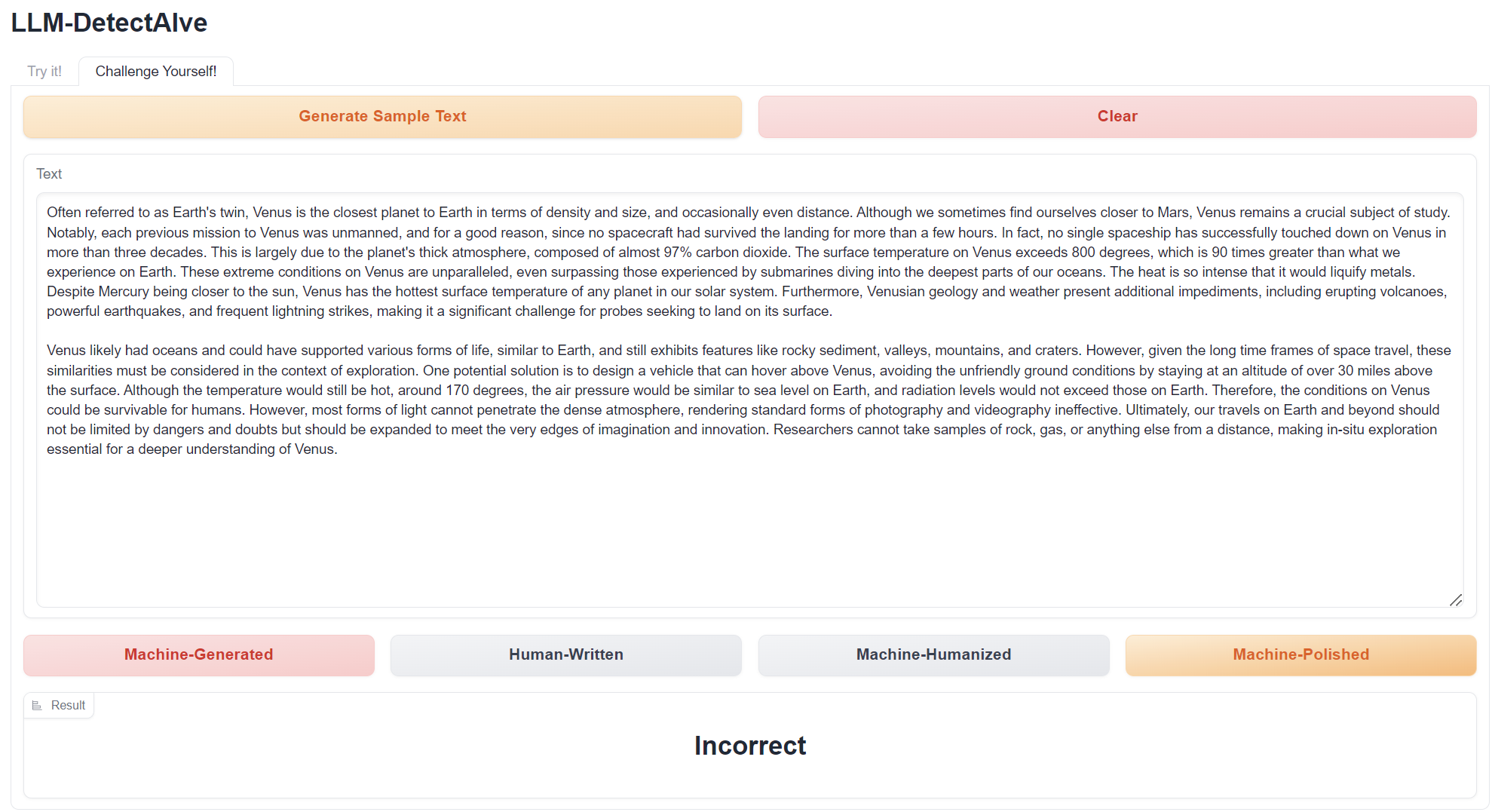} \label{fig:interface2}
\caption{\textbf{LLM-DetectAIve interface:} automatic text detection (top) and human detector playground (bottom).}
\label{fig:interface1}
\end{figure*}

Most prior work on detecting machine-generated text focused on binary detection, i.e.,~predicting whether the text is generated by a machine or written by a human. This dichotomy leaves no space for mixed categories of human-machine collaboration. However, we argue for the need for additional categories, as machine-polishing of human-written text is acceptable in certain cases (e.g., for academic papers), but not in other (e.g.,~in education).

In education, using LLMs to complete entire assignments or even to polish human-written essays is typically prohibited \citep{susnjak2022chatgptendonlineexam}.
Therefore, it is important to perform fine-grained text classification. 
For example, detecting the use of LLMs in text humanization and refinement becomes critical to ensure the fair assessment of students' genuine knowledge and abilities. Fine-grained human/machine identification is also important for authorship detection in digital forensics. A contemporary work by \citet{zhang-etal-2024-llm} also highlighted the importance of identifying LLM contributions as coauthors in a human-AI collaborative text.

To address this problem, we propose a new formulation of problem, as multi-way classification with the following labels:
\begin{itemize}[itemsep=1pt]
    \item[I.] \textbf{Human-Written}: the ext is created solely by a human author without GenAI assistance.

    \item[II.] \textbf{Machine-Generated}: the text is entirely produced by a machine based on input prompts without any human intervention.
    
    \item[III.] \textbf{Machine-Written Machine-Humanized}: the text is initially generated by a machine and then subtly modified to appear more human-like. This involves automatically tweaking the LLM to make the output appear more human.
    
    \item[IV.] \textbf{Human-Written Machine-Polished}: the text is written by a human and then is refined or polished by a machine, e.g.,~to correct grammar, improve style, and/or optimize readability while trying to preserve the meaning of the original human text.
    
\end{itemize} 

\begin{table*}[t!]
    \centering
    \resizebox{0.8\textwidth}{!}{
        \setlength{\tabcolsep}{3pt}
        \begin{tabular}{llccccccc}
            \toprule
            \textbf{Text Class} & \textbf{Generator} & \textbf{OUTFOX} & \textbf{Wikipedia} & \textbf{Wikihow} & \textbf{Reddit ELI5} & \textbf{arXiv abstract} & \textbf{PeerRead}\\
            \midrule
            \multicolumn{8}{c}{\textbf{M4GT-Bench}} \\
            \midrule
            \text{I} & \textbf{Human} &  14,043 & 14,333 & 15,999 & 16,000 & 15,998 & 2,847\\
            \midrule
            \multirow{6}{*}{II}  & \textbf{davinci-003} &  3,000 & 3,000 & 3,000 & 3,000 & 3,000 & 2,340\\
             & \textbf{gpt-3.5-turbo} &  3,000 & 2,995 & 3,000 & 3,000 & 3,000 & 2,340\\
             & \textbf{cohere} &  3,000 & 2,336 & 3,000 & 3,000 & 3,000 & 2,342\\
             & \textbf{dolly-v2} &  3,000 & 2,702 & 3,000 & 3,000 & 3,000 & 2,344\\
             & \textbf{BLOOMz} &  3,000 & 2,999 & 3,000 & 2,999 & 3,000 & 2,334\\
             & \textbf{gpt4} &  3,000 & 3,000 & 3,000 & 3,000 & 3,000 & 2,344\\
            \midrule
            \multicolumn{8}{c}{\textbf{New Generations}} \\
            \midrule
            \multirow{6}{*}{II + III + IV} & \textbf{gpt-4o} & 8,966 & 8,995 & 9,000 & 9,000 & 9,000 & 7,527\\
             & \textbf{gemma-7b} & 8,280 & 8,985 & 9,000 & 9,000 & 9,000 & 0\\
             & \textbf{llama3-8b}   & 8,271 & 8,985 & 9,000 & 9,000 & 9,000 & 0\\
             & \textbf{llama3-70b} & 8,577 & 8,985 & 9,000 & 9,000 & 9,000 & 0\\
             & \textbf{mixtral-8x7b} & 17,001 & 8,985 & 9,000 & 9,000 & 9,000 & 0\\
             & \textbf{gemma2-9b} &  0 & 8,985 & 9,000 & 9,000 & 9,000 & 0\\
            \midrule
            \multirow{2}{*}{III} & \textbf{gemini1.5} &  0 & 1,652 & 1,601 & 904 & 0 & 0\\
             & \textbf{mistral-7b} &  0 & 2,993 & 3,000 & 0 & 0 & 2,344\\
            \midrule
            \multirow{2}{*}{IV} & \textbf{gemini1.5} &  0 & 1,652 & 1,601 & 904 & 2,994 & 586\\
             & \textbf{mistral-7b} &  0 & 2,993 & 3,000 & 0 & 0 & 2,344\\
            \bottomrule
        \end{tabular}
    }
    \caption{\textbf{Statistics about our datasets} across LLMs over the four classes: I. \hw, II. \mg, III. \mwmh and IV. \hwmp. For row II + III + IV, the data is approximately uniformly distributed across the three classes.}
    \label{tab:dataset}
\end{table*}

We further develop \textbf{LLM-DetectAIve}, a system that accurately distinguishes between different types of text generation and editing. With this, we aim to uphold academic integrity and ensure a fair evaluation process for both students and researchers.

Our contributions are as follows:
\begin{itemize}[itemsep=1pt]
    \item We reformulate the task as fine-grained multi-way classification.
    \item We collect a dataset for this reformulation using generations from a variety of LLMs.
    \item We build, evaluate, and compare several machine-generated text detectors on our new fine-grained dataset.
    \item We develop a Web-based demo that (\emph{i})~allows users to input text and to obtain fine-grained classification prediction, and (\emph{ii})~offers a playground for users to test their 
    ability to detect texts with varying degrees of LLM involvement, according to the above 4-way fine-grained schema.  
\end{itemize}

\section{Dataset}
To collect the dataset for our multi-way fine-grained detector, we first gathered datasets that were curated for binary machine-generated text detection from previous work, and then we extended the data into our four labels by introducing new corresponding generations. Sections~\ref{sec:prompts} and \ref{sec:apitoolandcleaning} discuss the prompts we used for generation and data cleaning, respectively.

\subsection{Data Overview}

We build the new dataset by extending the M4GT-Bench \cite{wang2024m4gtbenchevaluationbenchmarkblackbox}, which is an benchmark dataset for evaluating machine-generation text detectors that encompasses multiple generators and domains, including arXiv, Wikihow, Wikipedia, Reddit, student essays (OUTFOX), and peer reviews (PeerRead). From these sources, we sampled a subset comprising 79,220 human-written texts and 103,075 machine-generated texts. 

Next, we expanded this dataset by (\emph{i})~collecting additional machine-generated texts produced by new LLMs (e.g., GPT-4o), (\emph{ii})~generating machine-written then machine-humanized texts, and (\emph{iii})~polishing human-written texts using various LLMs. 
This resulted in 91,358 fully-MGTs, 103,852 machine-written then machine-humanized texts, and 107,900 human-written then machine-polished texts. Table~\ref{tab:dataset} gives detailed statistics about the dataset.

For data generation, we used a variety of LLMs, including Llama3-8b, Llama3-70b~\cite{dubey2024llama3herdmodels}, Mixtral 8x7b \cite{jiang2024mixtralexperts}, Gemma-7b, Gemma2-9b \cite{gemmateam2024gemma2improvingopen}, GPT-4o~\citep{gpt4}, Gemini-1.5-pro~\citep{Gemini2023}, and Mistral-7b \cite{jiang2023mistral}. By incorporating a diverse array of LLMs and domains, we aim to enhance the detection accuracy within actual domains and generators, as well as improve generalization.

\subsection{Generation Prompts}
\label{sec:prompts}

For the \emph{Machine-Written Machine-Humanized} class, examples of prompts include \textit{Rewrite this text to make it sound more natural and human-written} or ``\textit{Rephrase this text to be easy to understand and personable.}'' For the \emph{Human-Written Machine-Polished} class, we used prompts such as ``\textit{Paraphrase the provided text.}'' or ``\textit{Rewrite this text so that it is grammatically correct and flows nicely.}'' Additionally, we introduced a trailing prompt appended to each randomly selected prompt to prevent undesirable text that the LLM may prepend to its output, e.g., ``\textit{Only output the text in double quotes with no text before or after it. Text: \{\} Your response:}''. 
We used 5-6 prompts per domain to generate data for the \emph{Machine-Written Machine-Humanized} and \emph{Human-Written Machine-Polished} classes. In addition to the \emph{Machine-Generated} class, we used the original prompts from the M4GT-Bench dataset.

\subsection{API Tools \& Data Cleaning}
\label{sec:apitoolandcleaning}
For data generation, we used multiple APIs from OpenAI, Gemini, Groq, and DeepInfra, to generate a total of 303,110 texts for the three LLM-dependent classes. For each of the two new class generations, we limited the text length to 1,500 words in order to accommodate the context length restrictions of some smaller LLMs and to efficiently manage time and costs. 

The output of the LLMs occasionally included formatting such as ``Here is the paraphrased text:'' and ``Sure!'' despite instructions in the trailing prompt to exclude any additional output. We removed these phrases in the post-processing with two considerations. On the one hand, this naturally occurs in real-world applications, i.e., humans will remove these irrelevant phrases when they use the target content.
Moreover, the presence of these indicative artifacts could impact the detectors' generalization and the quality of the dataset, given that they are potentially unique for a specific text class.

\section{Detection Models}



We trained three detectors by fine-tuning RoBERTa \citep{liu2019robertarobustlyoptimizedbert}, DeBERTa \citep{he2021deberta}, and DistilBERT \citep{Sanh2019DistilBERTAD}. 
DeBERTa is built upon BERT and RoBERTa by incorporating disentangled attention mechanisms and an enhanced mask decoder, which improves word representation. 

Eventually, in the demo, we used DistilBERT, which is a compact and fast variant of BERT: 60\% faster and 40\% smaller, while retaining 97\% of BERT's language understanding capabilities.

Table \ref{tab:hyperparameters} shows the values of the hyper-parameters for each model. We used RoBERTa and DistilBERT in our domain-specific experiments. However, due to the inferior performance of DistilBERT to RoBERTa in our preliminary trials, we substituted DistilBERT with DeBERTa in the following experiments (DeBERTa is superior to RoBERTa).

\begin{table}[t]
\centering
\resizebox{\columnwidth}{!}{%
\begin{tabular}{llcccc}
\toprule
\textbf{Dataset} & \textbf{Detector} & \textbf{Learning rate} & \textbf{Weight Decay} & \textbf{Epochs} & \textbf{Batch Size} \\
\midrule
\multirow{2}{*}{arXiv} & RoBERTa & 2e-5 & 0.01 & 10 & 16 \\
 & DistilBERT & 2e-5 & 0.01 & 10 & 16 \\
\midrule
\multirow{2}{*}{OUTFOX} & RoBERTa & 2e-5 & 0.01 & 10 & 16 \\
 & DistilBERT & 2e-5 & 0.01 & 10 & 16 \\
 \midrule
\multirow{2}{*}{Full Dataset} & RoBERTa & 5e-5 & 0.01 & 10 & 32 \\
 & DeBERTa & 5e-5 & 0.01 & 10 & 32\\
\bottomrule
\end{tabular}%
}
\caption{\textbf{Hyper-parameter values} across the models.}
\label{tab:hyperparameters}
\end{table}

\section{Experiments and Evaluation}
\label{sec:experiments}


The previous studies have shown that the accuracy of detectors drops substantially when testing on out-of-domain examples~\citep{wang2024m4gtbenchevaluationbenchmarkblackbox}.
To alleviate this, we propose three strategies: (\emph{i})~train multiple domain-specific detectors, each specifically responsible for detecting inputs from one domain, (\emph{ii})~train one universal detector using more training data across various domains, and (\emph{iii})~leverage domain-adversarial neural network (DANN) for domain adaption.

\subsection{Domain-Specific Detectors}
\label{sec:domainspecific}
\begin{figure}[t!]
    \centering
    \includegraphics[scale=0.4]{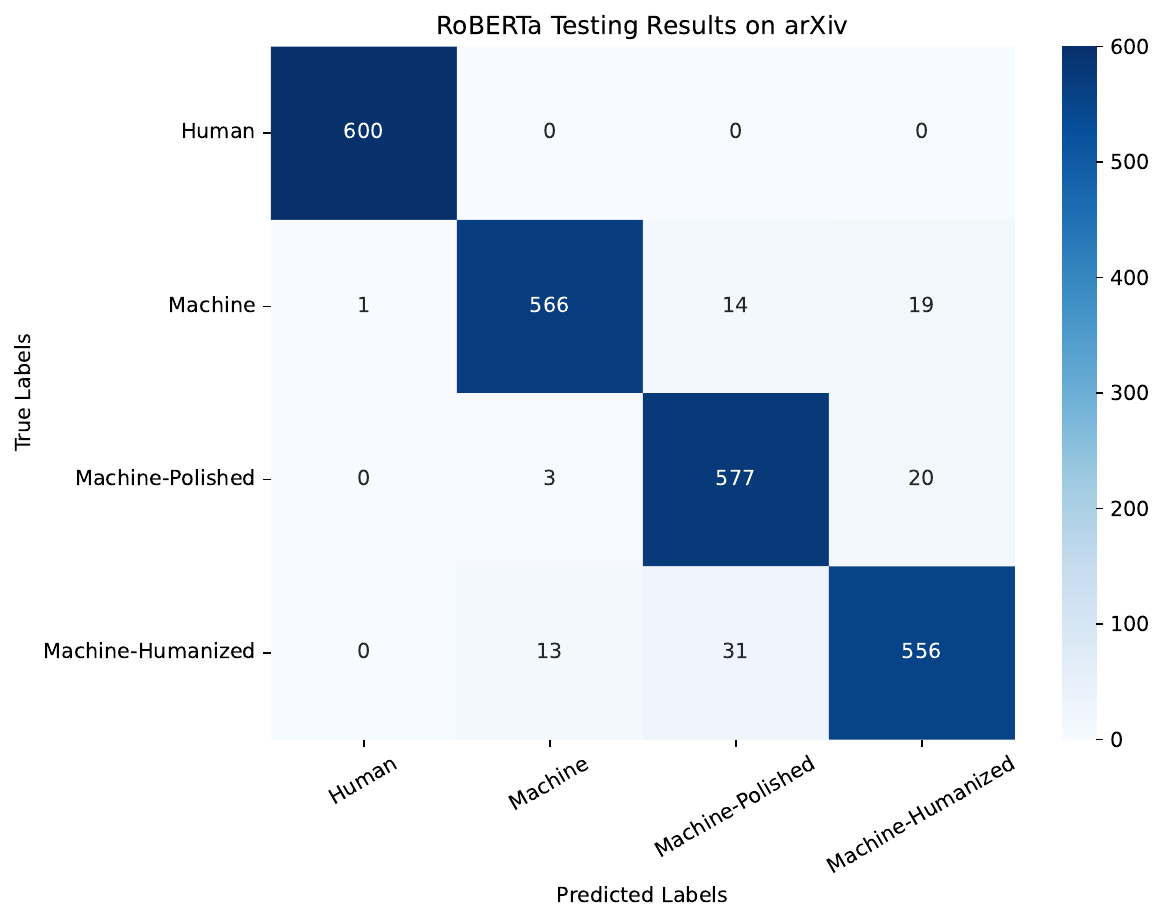}
    \includegraphics[scale=0.4]{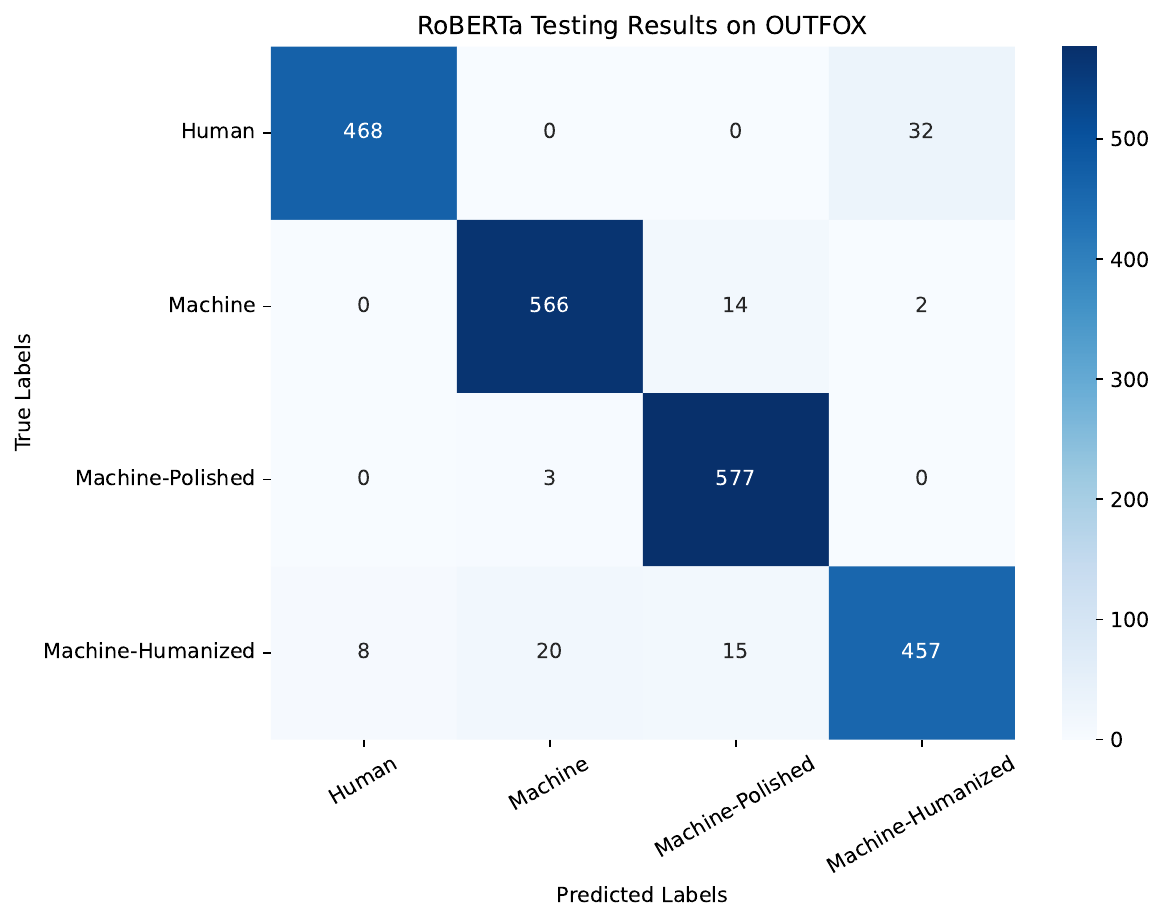}
\caption{Domain-specific confusion matrix for RoBERTa on arXiv (top) and on OUTFOX (bottom).}
\label{fig:confusion_matrices}
\end{figure}

We fine-tuned RoBERTa and DistilBERT using the data from arXiv and OUTFOX, using a ratio of training, validation, and test sets of 70\%:15\%:15\%. 
The results are shown in Table \ref{tab:performance}. We can see that both RoBERTa and DistillBERT performed well on OUTFOX. Overall, RoBERTa is more robust over diverse domains, with accuracy greater than 95\% on both domains, with a small number of mis-classifications occurring between classes with overlapping features, such as Machine-Generated vs. Human-Written, vs. Machine-Polished classes, as the confusion matrices in Figure~\ref{fig:confusion_matrices} show.

However, in this setup, the users need to first specify the domain of the input text, which is an extra effort. To mitigate this, we further trained a universal model that does not require the user to select the domain.

\subsection{Universal Detectors}
\label{sec:universal}
We fine-tuned RoBERTa and DeBERTa using the full dataset; the data distribution for this is shown in Table~\ref{tab:data_class_distribution}. 
To reduce data imbalance and prevent the detector from favoring any particular class, we excluded some of the original data. The evaluation results in Table~\ref{tab:full_dataset_performance} indicate that DeBERTa consistently outperforms RoBERTa across all evaluation measures we use. Therefore, we deployed the fine-tuned DeBERTa as the back\-end detection model for our demo.

\begin{table}[t]
\centering
\resizebox{\columnwidth}{!}{%
\begin{tabular}{llcccc}
\toprule
\textbf{Detector} & \textbf{Test Domain} & \textbf{Prec} & \textbf{Recall} & \textbf{F1-macro} & \textbf{Acc} \\
\midrule
\multirow{2}{*}{RoBERTa} & arXiv & 95.82 & 95.79 & 95.79 & 95.79 \\
 & OUTFOX & 95.67 & 95.43 & 95.53 & 95.65 \\
\midrule
\multirow{2}{*}{DistilBERT} & arXiv & 88.98 & 87.97 & 87.93 & 87.79 \\
 & OUTFOX & 96.66 & 96.65 & 96.65 & 96.65 \\
\bottomrule
\end{tabular}%
}
\caption{\textbf{Domain-specific performance} for RoBERTa and DistilBERT on arXiv and OUTFOX.}
\label{tab:performance}
\end{table}

\begin{table}[t]
    \centering
    \resizebox{\columnwidth}{!}{
        \begin{tabular}{lrrrr}
            \toprule
             & & \textbf{Machine-} & \textbf{Machine-} & \textbf{Machine-} \\
            \textbf{Domain} & \textbf{Human} & \textbf{Generated} & \textbf{Polished} & \textbf{Humanized} \\
            \midrule
            arXiv & 15,998 & 18,000 & 18,000 & 18,000 \\
            Reddit & 16,000 & 18,904 & 18,904 & 18,904 \\
            wikiHow & 15,999 & 22,601 & 22,601 & 22,601 \\
            Wikipedia & 14,333 & 22,615 & 22,615 & 22,615 \\
            PeerRead & 2,847 & 4,684 & 4,684 & 4,684 \\
            Outfox & 14,043 & 17,000 & 17,000 & 17,000 \\
            \bottomrule
        \end{tabular}
    }
    \caption{\textbf{Distribution} of the data used for fine-tuning \textbf{universal detectors} based on RoBERTa and DeBERTa.}
    \label{tab:data_class_distribution}
\end{table}

\begin{table}[th!]
    \centering
    \footnotesize
    \begin{tabular}{lcccc}
        \toprule
        \textbf{Detector} & \textbf{Prec} & \textbf{Recall} & \textbf{F1-Macro} & \textbf{Acc} \\
        \midrule
        RoBERTa & 94.79 & 94.63 & 94.65 & 94.62 \\
        DeBERTa & 95.71 & 95.78 & 95.72 & 95.71 \\
        \bottomrule
    \end{tabular}
    \caption{\textbf{Detector performance} on the full dataset.}
    \label{tab:full_dataset_performance}
\end{table}

\subsection{DANN-Based Detector}

In our domain-specific experiments above, we achieved strong performance when the domain of the text was provided. However, in cross-domain evaluation, the performance is sub-optimal as previous work has suggested~\citep{wang2024m4gtbenchevaluationbenchmarkblackbox, wang-etal-2024-m4}. In real-world scenarios, the domain would not always be specified, and thus we need a classifier that is as domain-independent as possible.. Thus, we investigated the use of \textit{domain adversarial neural networks}~\cite{dann} to train a domain-robust detector.

DANN was initially designed to achieve domain adaptation by aligning representations across different domains with three major components:
    \begin{itemize}
        \item \textbf{Representation Extractor:} which builds a representation of the input data; here, we use RoBERTa.
        \item \textbf{Label Predictor:} to predict the class labels based on the representation; it is trained using labeled data from the source domain.
        \item \textbf{Domain Classifier:} connected to the representation via a \textit{gradient reversal layer (GRL)}, it distinguishes between the source and the target domains. It multiplies the gradient by a negative constant during back-propagation, promoting domain-invariant representation.
    \end{itemize}

The network is trained using standard back-propagation and stochastic gradient descent, optimizing the label classification loss while intentionally confusing the model regarding the domain by reversing the gradient from the domain classifier. This reduces the label classification loss while increasing the domain classification one. 

As a result, the Domain-Adversarial Neural Network (DANN) yields a representation that is independent of the domain. In our experiments, we trained the DANN to predict our four classes and to be as confused as possible when predicting the six sources/domains.
The results are shown in Table~\ref{tab:dann_performance}. We can see that using domain adversarial training on top of RoBERTa-enhances the overall performance compared to just fine-tuning RoBERTa as in Section \ref{sec:universal}. This suggests that decoupling the model from domain-specific representation leads to an improvement in its overall performance.

\begin{table}[t!]
    \centering
    \footnotesize
    \resizebox{\columnwidth}{!}{
    \begin{tabular}{lcccc}
        \toprule
        \textbf{Detector} & \textbf{Prec} & \textbf{Recall} & \textbf{F1-macro} & \textbf{Acc} \\
        \midrule
        RoBERTa & 94.79 & 94.63 & 94.65 & 94.62 \\
        DANN+RoBERTa & \textbf{96.30} & \textbf{95.54} & \textbf{96.06} & \textbf{95.24} \\
        \bottomrule
    \end{tabular}
    }
    \caption{Comparing domain-specific RoBERTa vs. DANN+RoBERTa. The latter outperforms the former across all measures, indicating that decoupling the model from domain-specific representation is beneficial.}
    \label{tab:dann_performance}
\end{table}

\subsection{Comparison to Existing Systems}
There are several previously proposed systems for detecting machine-generated text, such as GPTZero,\footnote{\url{https://gptzero.me/}} ZeroGPT,\footnote{\url{ https://www.zerogpt.com/}} and Sapling AI detector,\footnote{\url{https://sapling.ai/ai-content-detector}} but none of them supports four classes. GPTZero is the only one that goes beyond binary classification: it adds a \emph{mixed text}; however, it limits users to only 40 free runs per day or 10,000 words per month for registered accounts. Thus, we could not perform comparison on our entire test dataset. Instead, we randomly sampled 60 machine-generated texts and 60 human texts (10 per source) per source. In this binary classification setting, LLM-DetectAIve achieved 97.50\% accuracy, outperforming GPTZero, ZeroGPT, and Sapling AI, with 87.50\%, 69.17\%, and 88.33\%, respectively.

\subsection{Generalization Evaluation}
To evaluate the generalization ability of our detector on unseen domains and generators, we experimented with testing on two additional datasets: MixSet~\citep{zhang-etal-2024-llm} and IELTS essays written by individuals for whom English is a second language.\footnote{\url{https://huggingface.co/datasets/chillies/IELTS_essay_human_feedback}} 

For the IELTS essays, after deduplication, we randomly sampled 300 (essay problem statement, human-written essay) pairs, and then we produced the corresponding machine-written essays using the problem statements based on Llama3.1-70B. We further generated \mwmh and \hwmp. 
For MixSet, the original dataset contains a total of 3,600 examples, with 300, 300, 600, and 2,400 examples for \hw, \mg, \mwmh and \hwmp, respectively.
It involves models such as Llama2-70B and GPT-4, and text covering domains of email content, news, game reviews, and so on. 

The results are shown in Table~\ref{tab:unseen_performance}, where we can see that the detector performs much worse on unseen domains and generators, compared to in-domain and in-generator cases. The performance on IELTS is better than on MixSet.
This can be attributed to the inclusion of the OUTFOX data (English native-speaker student essays) in the training data, while the domains and the generators in MixSet are not in the training set. 
The low generalization performance suggests challenges in adapting black-box detectors to the diverse domains and generators in real-world applications.

\begin{table}[t!]
    \centering
    \footnotesize
    \resizebox{0.84\columnwidth}{!}{
    \begin{tabular}{lcccc}
        \toprule
        \textbf{Dataset} & \textbf{Prec} & \textbf{Recall} & \textbf{F1-macro} & \textbf{Acc} \\
        \midrule
        IELTS & 63.74 & 66.91 & 66.55 & 66.91 \\
        MixSet & 59.18 & 64.25 & 54.95 & 60.08 \\
        \bottomrule
    \end{tabular}
    }
    \caption{\textbf{Cross-domain evaluation} of our detector on unseen domains and generators: IELTS and MixSet.}
    \label{tab:unseen_performance}
\end{table}

\section{Demo Web Application}

Our demo web application has two interfaces: (\emph{i})~an interface for fine-grained MGT detection, and (\emph{ii})~a playground for users.

\subsection{Automatic Detection}
The automatic detection interface is shown in Figure~\ref{fig:interface1} (top). It allows users to input a text, and then the system responds with the class that the text belongs to. To ensure the prediction accuracy, the length of the submitted text is constrained to 50-500 words since the performance of our detectors drops significantly for shorter texts.
Longer texts will be truncated, as we are limited by the context window size of the BERT-like transformers we use.

\subsection{Human Detector Playground}
The demo further offers a human detector playground as an interactive interface, which allows users to test their capability to distinguish between the four text categories. Figure~\ref{fig:interface1} (bottom) shows a snapshot of the playground interface where the users can try the system, gaining insights into the subtle differences between various types of human-written and machine-generated texts.

\subsection{Deployment and Implementation}
Our demo is deployed on Hugging Face Spaces, which allows seamless integration with transformer models, ease of use, and robust support for hosting machine learning applications. For implementing the user interface, we used Gradio. The code is publicly available under an MIT license.

\section{Conclusion and Future Work} 
In an era of advanced large language models, maintaining the integrity of text poses significant challenges. We presented a system that aims to identify the use of machine-generated text, accurately differentiating human-written text from various types of automatically generated text. 
Unlike previous work, we use a fine-grained classification schema (Human-Written, Machine-Generated, Machine-Written Machine-Humanized, and Human-Written Machine-Polished), which offers insights into the origins of the text, thus enabling trustworthiness.

In future work, we plan to improve the Domain Adversarial Neural Network (DANN) to improve the results even further. We further plan to explore the possibility of using a DANN on the text's generator instead of the text's domain to generalize detection across different text generators. Using a DANN on both the domain and the generator could potentially lead to a truly universal detector. We also aim to expand the classification to include a fifth category: machine-written and human-edited text, enhancing detection capabilities and providing a more comprehensive analysis of text origins. To further improve the system, we also plan to address potential biases in the dataset caused by formatting styles linked to specific domains, such as Wikihow and PeerRead, to ensure better robustness across a broader range of human-written content. Last but not least, we want to expand the dataset to encompass a diverse set of languages, enabling the development of a robust multilingual detection model.

\section*{Limitations}

We acknowledge certain limitations of our work, which we plan to address in future work. First, although our work has explored more fine-grained machine-generated text scenarios beyond conventional binary classification, we did not consider a complex scenario where the text is first generated by a machine and then is manually edited by humans to suit their personal needs. This is primarily due to the high costs associated with collecting data that requires human editing.

Moreover, we identified some issues with the dataset. Specifically, some LLMs associate specific domains with particular formatting styles, such as markdown for lists, bullet points, and headers. This issue was particularly noticeable in the Wikihow and PeerRead domains, where the LLMs frequently applied these formatting styles, potentially skewing the data and impacting the accuracy of our classifications. It also remains uncertain whether our system can generalize to detecting models or languages not included in our English-only dataset.

\section*{Ethical Statement and Broad Impact}

\paragraph{Data License} 
A primary ethical consideration is the data license. We reused pre-existing corpora, such as OUTFOX and Wikipedia, which have been publicly released and approved for research purposes. Moreover, we generated new data on top of the original data, thereby mitigating concerns regarding data licensing.

\paragraph{Biased and Offensive Language} 
Considering that our data is generated by large language models, it might contain offensive or biased language; we did not try to control for this, replying on the inetnal safety mechanisms of the LLMs we used.

\paragraph{Positive Impact of Fine-Grained Detection} 
LLM-DetectAIve expands the conventional binary classification in machine-generated text detection to more fine-grained levels, which is more aligned with real-life scenarios. We believe this approach could be applied in various scenarios, e.g.,~for students' essays to ensure the originality of their work. Moreover, LLM usage detection may find applications in authorship detection as well as in digital forensics.

\bibliography{custom}
\bibliographystyle{acl_natbib}

\include{emnlp2024-demo/section/7_appendix}

\end{document}